# Cluster-based Specification Techniques in Dempster-Shafer Theory*

Johan Schubert

Division of Information System Technology,
Department of Command and Control Warfare Technology,
National Defence Research Establishment,
S-172 90  Stockholm, SWEDEN
E-mail: schubert@sto.foa.se

**Abstract.** When reasoning with uncertainty there are many situations where evidences are not only uncertain but their propositions may also be weakly specified in the sense that it may not be certain to which event a proposition is referring. It is then crucial not to combine such evidences in the mistaken belief that they are referring to the same event. This situation would become manageable if the evidences could be clustered into subsets representing events that should be handled separately. In an earlier article we established within Dempster-Shafer theory a criterion function called the metaconflict function. With this criterion we can partition a set of evidences into subsets. Each subset representing a separate event. In this article we will not only find the most plausible subset for each piece of evidence, we will also find the plausibility for every subset that the evidence belongs to the subset. Also, when the number of subsets are uncertain we aim to find a posterior probability distribution regarding the number of subsets.

## 1   Introduction

In an earlier article [1] we derived a method, within the framework of Dempster-Shafer theory [2-3], to handle evidences that are weakly specified in the sense that it may not be certain to which of several possible events a proposition is referring. When reasoning with such evidences we must avoid combining evidences by mistake that refer to different events. The situation would become manageable if the evidences could be clustered into subsets representing events that should be handled separately. For this reason every proposition's action part must be supplemented with an event part describing to which event the proposition is referring. The event part may be more or less weakly specified dependent on the evidence.

An example from our earlier article illustrates the terminology:

> Let us consider the burglaries of two bakers' shops at One and Two Baker Street, event 1 ($E_1$) and event 2 ($E_2$), i.e., the number of events is known to be two. One witness hands over an evidence, specific with respect to event, with the proposition: "The burglar at One Baker Street," event part: $E_1$, "was probably brown haired (B)," action part: B. A second anonymous witness hands over a nonspecific evidence with the proposition: "The burglar at Baker Street," event part: $E_1$, $E_2$, "might have been red haired (R)," action part: R. That is, for example:

---

*This paper is based on my Ph.D. thesis [6].

*evidence* 1:
  *proposition*:
    *action part: B*
    *event part: $E_1$*
  $m(B) = 0.8$
  $m(\Theta) = 0.2$

*evidence* 2:
  *proposition*:
    *action part: R*
    *event part: $E_1, E_2$*
  $m(R) = 0.4$
  $m(\Theta) = 0.6$

In this situation it is impossible to directly separate evidences based only on their proposition. Instead we will use the conflict between the propositions of two evidences as a probability that the two reports are referring to different events.

The general idea is this. If we receive evidences about several different and separate events and the evidences are mixed up, we want to sort the evidences according to which event they are referring to. Thus, we partition the set of all evidences $\chi$ into subsets where each subset refers to a particular event. In Figure 1 these subsets are denoted by $\chi_i$ and the conflict in $\chi_i$ is denoted by $c_i$. Here, thirteen evidences are partitioned into four subsets. When the number of subsets is uncertain there will also be a "domain conflict" $c_0$ which is a conflict between the current number of subsets and domain knowledge. The partition is then simply an allocation of all evidences to the different events. Since these events do not have anything to do with each other, we will analyze them separately.

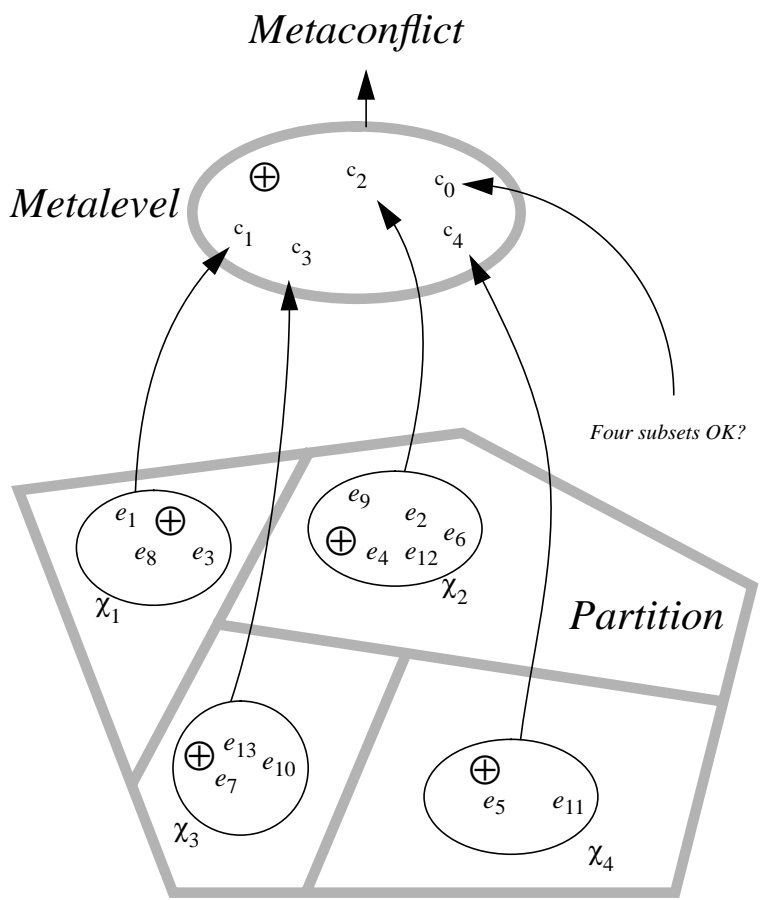

**Fig. 1.** The Conflict in each subset of the partition becomes an evidence at the metalevel.

Now, if it is uncertain to which event some evidence is referring we have a problem. It could then be impossible to know directly if two different evidences are referring to the same event. We do not know if we should put them into the same subset or not. This problem is then a problem of organization. Evidences from different problems that we want to analyze are unfortunately mixed up and we are having some problem separating them.

To solve this problem, we can use the conflict in Dempster's rule when all evidences within a subset are combined, as an indication of whether these evidences belong together. The higher this conflict is, the less credible that they belong together.

Let us create an additional piece of evidence for each subset where the proposition of this additional evidence states that this is not an "adequate partition". Let the proposition take a value equal to the conflict of the combination within the subset. These new evidences, one regarding each subset, reason about the partition of the original evidences. Just so we do not confuse them with the original evidences, let us call these evidences "metalevel evidences" and let us say that their combination and the analysis of that combination take place on the "metalevel", Figure 1.

In the combination of all metalevel evidences we only receive support stating that this is not an "adequate partition". We may call this support a "metaconflict". The smaller this support is, the more credible the partition. Thus, the most credible partition is the one that minimizes the metaconflict.

This methodology was intended for a multiple-target tracking algorithm in an anti-submarine intelligence analysis system [4]. In this application a sparse flow of intelligence reports arrives at the analysis system. These reports may originate from several different unconnected sensor systems. The reports carry a proposition about the occurrence of a submarine at a specified time and place, a probability of the truthfulness of the report and may contain additional information such as velocity, direction and type of submarine.

When there are several submarines we want to separate the intelligence reports into subsets according to which submarine they are referring to. We will then analyze the reports for each submarine separately. However, the intelligence reports are never labeled as to which submarine they are referring to. Thus, it is not possible to directly differentiate between two different submarines using two intelligence reports.

Instead we will use the conflict between the propositions of two intelligence reports as a probability that the two reports are referring to different submarines. This probability is the basis for separating intelligence reports into subsets.

The cause of the conflict can be non-firing sensors placed between the positions of the two reports, the required velocity to travel between the positions of the two reports at their respective times in relation to the assumed velocity of the submarines, etc.

## 2 Separating Nonspecific Evidence

In [1] we established a criterion function of overall conflict called the metaconflict function. With this criterion we can partition evidences with weakly specified propositions into subsets, each subset representing a separate event. We will use the minimizing of the metaconflict function as the method of partitioning the set of evidences into subsets. This method will also handle the situation when the number of events are uncertain.

An algorithm for minimizing the overall conflict was proposed. The proposed algorithm is based on the one hand on characteristics of the criterion function for varying number of subsets and on the other hand on an iterative optimization among partitionings of evidence for a fixed number of subsets.

The conflict in Dempster's rule measures the lack of compatibility between evidences. Since evidences referring to different events tend to be more incompatible than evidences referring to the same event, it is an obvious choice as a distance measure between evidences in a cluster algorithm.

We have a conflict between two pieces of evidence within the same subset in two different situations. First, we have a conflict if the proposition action parts are conflicting regardless of the proposition event parts since they are presumed to be referring to the same event. Secondly, if the proposition event parts are conflicting then, regardless of the proposition action parts, we have a conflict with the presumption that they are referring to the same event. The idea of using the conflict in Dempster's rule as distance measure between evidences was first suggested by Lowrance and Garvey [5].

The metaconflict used to partition the set of evidences is derived as the plausibility that the partitioning is correct when the conflict in each subset is viewed as a metalevel evidence against the partitioning of the set of evidences, $\chi$, into the subsets, $\chi_i$. We have a simple frame of discernment on the metalevel $\Theta = \{\text{AdP}, \neg\text{AdP}\}$, where AdP is short for "adequate partition", and a basic probability assignment (bpa) from each subset $\chi_i$ assigning support to a proposition against the partitioning:

$$m_{\chi_i}(\neg\text{AdP}) \stackrel{\Delta}{=} \text{Conf}(\{e_j | e_j \in \chi_i\}),$$

$$m_{\chi_i}(\Theta) \stackrel{\Delta}{=} 1 - \text{Conf}(\{e_i | e_i \in \chi_i\})$$

where $e_j$ is the $j$th evidence and $\{e_j | e_j \in \chi_i\}$ is the set of evidences belonging to subset $\chi_i$ and $\text{Conf}(\cdot)$ is the conflict, $k$, in Dempster's rule. Also, we have a bpa concerning the domain resulting from a probability distribution about the number of subsets, $E$, conflicting with the actual current number of subsets, $\#\chi$. This bpa also assigns support to a proposition against the partitioning:

$$m_D(\neg\text{AdP}) \stackrel{\Delta}{=} \text{Conf}(\{E, \#\chi\}),$$

$$m_D(\Theta) \stackrel{\Delta}{=} 1 - \text{Conf}(\{E, \#\chi\}).$$

The combination of these by Dempster's rule give us the following plausibility of the partitioning:

$$\text{Pls}(\text{AdP}) = (1 - m_D(\neg\text{AdP})) \cdot \prod_{i=1}^{r} \left(1 - m_{\chi_i}(\neg\text{AdP})\right).$$

Finding the most probable partitioning of evidences into disjoint subsets representing different events will then be the problem of maximizing the plausibility of possible partitionings, or the dual problem of minimizing one minus the plausibility. The difference, one minus the plausibility of a partitioning, will be called the metaconflict of the partitioning.

## 2.1 Metaconflict as a Criterion Function

Let $E_i$ be a proposition that there are $i$ subsets, $\Theta_E = \{E_0, ..., E_n\}$ a frame of domain propositions and $m(E_i)$ the support for proposition $E_i$.

The metaconflict function can then be defined as:

DEFINITION. *Let the* metaconflict function,

$$Mcf(r, e_1, e_2, ..., e_n) \triangleq 1 - (1 - c_0) \cdot \prod_{i=1}^{r} (1 - c_i),$$

*be the conflict against a partitioning of n evidences of the set* $\chi$ *into r disjoint subsets* $\chi_i$. *Here, $c_i$ is the conflict in subset i and $c_0$,*

$$c_0 = \sum_{i \neq r} m(E_i),$$

*is the conflict between r subsets and propositions about possible different number of subsets.*

Two theorems are derived to be used in the separation of the set of evidences into subsets by an iterative minimization of the metaconflict function. By using these theorems we are able to reason about the optimal estimate of the number of events, when the actual number of events may be uncertain, as well as the optimal partition of nonspecific evidence for any fixed number of events. These two theorems will also be useful in a process for specifying evidences by observing changes in the metaconflict when moving a single piece of evidences between different subsets.

THEOREM 1. *For all j with $j < r$, if $m(E_j) < m(E_r)$ then min $Mcf(r,e_1,e_2,...,e_n) <$ min $Mcf(j,e_1,e_2,...,e_n)$.*

This theorem states that an optimal partitioning for $r$ subsets is always better than the other solutions with fewer than $r$ subsets if the basic probability assignment for $r$ subsets is greater than the basic probability assignment for the fewer subsets.

THEOREM 2. *For all j, if min $Mcf(r, e_1, e_2, ..., e_n) < \sum_{i \neq j} m(E_i)$ then min $Mcf(r,e_1,e_2,...,e_n) <$ min $Mcf(j,e_1,e_2,...,e_n)$.*

Theorem 2 states that an optimal partitioning for some number of subsets is always better than other solutions for any other number of subsets when the domain part of the metaconflict function is greater than the total metaconflict of the present partitioning.

For a fixed number of subsets a minimum of the metaconflict function can be found by an iterative optimization among partitionings of evidences into different subsets. This approach is proposed in order to avoid the combinatorial problem in minimizing the metaconflict function. In each step of the optimization the consequence of transferring an evidence from one subset to another is investigated.

The algorithm for finding the partitioning of evidences among subsets that minimizes the metaconflict is based on theorems 1 and 2 of the metaconflict function for finding the optimal number of subsets and an iterative optimization among partitionings of evidences for a fixed number of subsets. The iterative part of the algorithm guarantees, like all hill climbing algorithms, local but not global optimum.

## 3  Specifying Nonspecific Evidence

### 3.1  Evidences About Evidence

A conflict in a subset $\chi_i$ is interpreted as an evidence that there is at least one piece of evidence that does not belong to the subset;

$$m_{\chi_i}(\exists j . e_j \notin \chi_i) = c_i.$$

If an evidence $e_q$ in $\chi_i$ is taken out from the subset the conflict $c_i$ in $\chi_i$ decreases to $c_i^*$. This decrease $c_i - c_i^*$ is interpreted as an evidence indicating that $e_q$ does not belong to $\chi_i$, $m_{\Delta\chi_i}(e_q \notin \chi_i)$, and the remaining conflict $c_i^*$ is an other evidence indicating that there is at least one other evidence $e_j$, $j \neq q$, that does not belong to $\chi_i - \{e_q\}$,

$$m_{\chi_i - \{e_q\}}(\exists j \neq q . e_j \notin (\chi_i - \{e_q\})) = c_i^*.$$

The unknown bpa, $m_{\Delta\chi_i}(e_q \notin \chi_i)$, is derived by stating that the belief that there is at least one piece of evidence that does not belong to $\chi_i$ should be equal, no matter whether that belief is based on the original evidence $m_{\chi_i}(\exists j . e_j \notin \chi_i)$, before $e_q$ is taken out from $\chi_i$, or on a combination of the other two evidences $m_{\Delta\chi_i}(e_q \notin \chi_i)$ and $m_{\chi_i - \{e_q\}}(\exists j \neq q . e_j \notin (\chi_i - \{e_q\}))$, after $e_q$ is taken out from $\chi_i$, i.e.

$$\text{Bel}_{\chi_i}(\exists j . e_j \notin \chi_i) = \text{Bel}_{\Delta\chi_i \oplus (\chi_i - \{e_q\})}(\exists j . e_j \notin \chi_i).$$

where

$$\text{Bel}_{\chi_i}(\exists j . e_j \notin \chi_i) = c_i$$

and

$$\text{Bel}_{\Delta\chi_i \oplus (\chi_i - \{e_q\})}(\exists j . e_j \notin \chi_i) = c_i^* + m_{\Delta\chi_i}(e_q \notin \chi_i) \cdot [1 - c_i^*].$$

Thus, we have derived the evidence that $e_q$ does not belong to $\chi_i$ from the variations in cluster conflict when $e_q$ was taken out from $\chi_i$:

$$m_{\Delta\chi_i}(e_q \notin \chi_i) = \frac{c_i - c_i^*}{1 - c_i^*}.$$

If $e_q$ after it is taken out from $\chi_i$ is brought into another subset $\chi_k$, its conflict will increase from $c_k$ to $c_k^*$. The increase in conflict when $e_q$ is brought into $\chi_k$ is interpreted as if there exists some evidence indicating that $e_q$ does not belong to $\chi_k + \{e_q\}$, i.e.

$$\forall k \neq i . m_{\Delta \chi_k}(e_q \notin (\chi_k + \{e_q\})) = \frac{c_k^* - c_k}{1 - c_k}.$$

When we take out an evidence $e_q$ from subset $\chi_i$ and move it to some other subset we might have a changes in domain conflict. The domain conflict is interpreted as an evidence that there exists at least one piece of evidence that does not belong to any of the $n$ first subsets, $n = |\chi|$, or if that particular evidence was in a subset by itself, as an evidence that it belongs to one of the other $n$-1 subsets. This indicate that the number of subsets is incorrect.

When $|\chi_i| > 1$ we may not only put an evidence $e_q$ that we have taken out from $\chi_i$ into another already existing subset, we may also put $e_q$ into a new subset $\chi_{n+1}$ by itself. There is no change in the domain conflict when we take out $e_q$ from $\chi_i$ since $|\chi_i| > 1$. However, we will get an increase in domain conflict from $c_0$ to $c_0^*$ when we move $e_q$ to $\chi_{n+1}$. This increase is an evidence indicating that $e_q$ does not belong to $\chi_{n+1}$, i.e.

$$m_{\Delta \chi}(e_q \notin \chi_{n+1}) = \frac{c_0^* - c_0}{1 - c_0}.$$

We will also receive an evidence from domain conflict variations if $e_q$ is in a subset $\chi_i$ by itself and moved from $\chi_i$ to another already existing subset. In this case we may get either an increase or decrease in domain conflict. First, if the domain conflict decreases $c_0^* < c_0$ when we move $e_q$ out from $\chi_i$ this is interpreted as an evidence that $e_q$ does not belongs to $\chi_i$,

$$m_{\Delta \chi}(e_q \notin \chi_i) = \frac{c_0 - c_0^*}{1 - c_0^*}.$$

Secondly, if we observe an increase in domain conflict $c_0^* > c_0$ we will interpret this as a new type of evidence, supporting the case that $e_q$ does belong to $\chi_i$;

$$m_{\Delta \chi}(e_q \in \chi_i) = \frac{c_0}{c_0^*}.$$

### 3.2 Specifying Evidences

We may now make a partial specification of each piece of evidence. We combine all evidence from different subsets regarding a particular piece of evidence and calculate for each subset the belief and plausibility that this piece of evidence belongs to the subset. The belief in this will always be zero, with one exception, since every proposition states that our evidence does not belong to some subset. The exception is when our evidence is in a subset by itself and we receive an increase in domain conflict when it is moved to an other subset. That was interpreted as if there exists an evidence that our piece of evidence does belong to the subset where it is placed. We will then also have a nonzero belief in that our piece of evidence belongs to the subset.

For the case when $e_q$ is in $\chi_i$ and $|\chi_i| > 1$ we receive, for example,

$$\forall k \neq n+1 . \text{Pls}(e_q \in \chi_k) = \frac{1 - m(e_q \notin \chi_k)}{1 - \prod_{j=1}^{n+1} m(e_q \notin \chi_j)}.$$

In the combination of all evidences regarding our piece of evidence we may receive support for a proposition stating that it does not belong to any of the subsets and can not be put into a subset by itself. That proposition is false and its support is the conflict in Dempster's rule, and also an indication that the evidence might be false.

In a subsequent reasoning process we will discount evidences based on their degree of falsity. If we had no indication as to the possible falsity of the evidence we would take no action, but if there existed such an indication we would pay ever less regard to the evidence the higher the degree was that the evidence is false and pay no attention to the evidence when it is certainly false. This is done by discounting the evidence with one minus the support of the false proposition.

Also, it is apparent that some evidences, due to a partial specification of affiliation, might belong to one of several different subsets. Such a piece of evidence is not so useful and should not be allowed to strongly influence the subsequent reasoning process within a subset.

If we plan to use an evidence in the reasoning process of some subset, we must find a credibility that it belongs to the subset in question. An evidence that cannot possible belong to a subset has a credibility of zero and should be discounted entirely for that subset, while an evidence which cannot possibly belong to any other subset and is without any support whatsoever against this subset has a credibility of one and should not be discounted at all when used in the reasoning process for this subset. That is, the degree to which an evidence can belong to a subset and no other subset corresponds to the importance the evidence should be allowed to play in that subset.

Here we should note that each original piece of evidence regardless of in which subset it was placed can be used in the reasoning process of any subset that it belongs to with a plausibility above zero, given only that it is discounted to its credibility in belonging to the subset.

When we begin our subsequent reasoning process in each subset, it will naturally be of vital importance to know to which event the subset is referring. This information is obtainable when the evidences in the subset have been combined. After the combination, each focal element of the final bpa will in addition to supporting some proposition regarding an action also be referring to one or more events where the proposed action may have taken place. Instead of summing up support for each event and every subset separately, we bring the problem to the metalevel where we simultaneously reason about all subsets, i.e. which subsets are referring to which events. In this analysis we use our domain knowledge stating that no more than one subset may be referring to an event. From each subset we then have an evidence indicating which events it might be referring to. We combining all the evidence from all different subsets with the restriction that any intersection in the combination that assigns one event to two different subsets is false. This method has a much higher chance to give a clearly preferable answer regarding which events is represented by which subsets, than that of only viewing the evidences within a subset when trying to determine its event.

The extension in this article of the methodology to partition nonspecific evidence developed in the first article [1] imply that an evidence will now be handled similarly by the subsequent reasoning process in different subsets if these are of approximately equal plausibility for the evidence. Without this extension the most plausible subset would take the evidence as certainly belonging to the subset while the other subsets would never consider the evidence at all in their reasoning processes.

## 4   Deriving a Posterior Domain Probability Distribution

Here we aim to find a posterior probability distribution regarding the number of subsets by combining a given prior distribution with evidence regarding the number of subsets that we received from the evidence specifying process.

We use the idea that each single piece of evidence in a subset supports the existence of that subset to the degree that this evidence supports anything at all other than the entire frame. In the evidence specifying process of the previous article we discounted each single evidence $m_q$ for its degree of falsity and its degree of credibility in belonging to the subset where it was placed, $\tilde{\tilde{m}}_q$. For each subset separately, we now combine all evidence within a subset and the resulting evidence is the total support for that subset. Thus we have

$$m_{\chi_i}(\chi_i \in \chi) = 1 - \frac{1}{1-k} \cdot \prod_q \tilde{\tilde{m}}_q(\Theta).$$

The degree to which the resulting evidence from this combination in its turn supports anything at all other than the entire frame, is then the degree to which all the evidence within the subset taken together supports the existence of this subset, i.e. that it is a nonempty subset that belongs to set of all subsets.

For every original piece of evidence we derived in the previous section an evidence with support for a proposition stating that this piece of evidence does not belong to the subset. If we have such support for every single piece of evidence in some subset, then this is also support that the subset is false. In that case none of the evidences that could belong to the subset actually did so and the subset was derived by mistake. Thus, we will discount the just derived evidence that support the existence of the subsets for this possibility.

Such discounted evidences $\tilde{m}_{\chi_i}$ that support the existence of different subsets, one from each subset, are then combined. The resulting bpa $\tilde{m}_\chi$ will have focal elements that are conjunctions of terms. Each term give support in that some particular subset belongs the set of all subsets, i.e. that it is a nonempty subset.

From this we can create a new bpa that is concerned with the question of how many subsets we have. This is done by exchanging each and every proposition in the previous bpa that is a conjunction of $r$ terms for one proposition in the new bpa that is on the form $|\chi| \geq r$, where $\chi$ is the set of all subsets. The sum of support of all focal elements in the previous bpa that are conjunctions of length $r$ is then awarded the focal element in the new bpa which supports the proposition that $|\chi| \geq r$;

$$m_\chi(|\chi| \geq r) = \sum_{\chi^* | |\chi^*| = r} m_\chi^{\%}((\ \wedge \chi^*) \in \chi),$$

where $\chi^* \in 2^\chi$ and $\chi = \{\chi_1, \chi_2, ..., \chi_n\}$.

A proposition in the new bpa is then a statement about the existence of a minimal number of subsets. Thus, where the previous bpa is concerned with the question of which subsets have support, the new bpa is concerned with the question of how many subsets are supported. This new bpa gives us some opinion that is based only on the evidence specifying process, about the probability of different numbers of subsets.

In order to obtain the sought-after posterior domain probability distribution we combine this newly created bpa that is concerned with the number of subsets with our prior domain probability distribution which was given to us in the problem specification.

Thus, by viewing each evidence in a subset as support for the existence of that subset we were able to derive a bpa, concerned with the question of how many subsets we have, which we could combine with our prior domain probability distribution in order to obtain the sought-after posterior domain probability distribution.

## 5 Conclusions

In this paper we have extended the methodology to partition nonspecific evidence developed in our previous article [1] to a methodology for specifying nonspecific evidence. This is in itself clearly an important extension in analysis, considering that an evidence will now in a subsequent reasoning process be handled similarly by different subsets if these are approximately equally plausible, whereas before the most plausible subset would take the evidence as certainly belonging to the subset while the other subsets would never consider the evidence in their reasoning processes.

We have also shown that it is possible to derive a posterior domain probability distribution from the reasoning process of specifying nonspecific evidence.